# Beyond Large Language Models: A Mental World Model Framework for Mental Modeling, Simulation, Evolution, and Interpretability

Qingshi Wang

*School of Computer Science, University of Bristol, Bristol, United Kingdom*
wangqingshi2004@gmail.com

Huiqi Tong

*Department of Psychiatry and Behavioral Sciences, Stanford University School of Medicine, Stanford, CA, United States*
htong@stanford.edu

Abstract: Understanding and simulating how subjective cognition and mental states emerge, evolve, and shape behavior is essential for advancing artificial intelligence toward social cognition and general intelligence. Existing large language models (LLMs) and Theory of Mind (ToM) methods primarily infer mental states from linguistic information, lacking unified mechanisms for subjective world construction, cognitive–psychological interactions, dynamic evolution, and historical dependencies. This paper introduces the Mental World Model (MWM) and its implementation, the Multi-Branch Mental Evolution Architecture (MMEA), which constructs subjective worlds, models interactions between cognition and mental states, and simulates mental evolution shaped by historical experiences. On DynToM and RecToM, MMEA achieves 96.48% and 94.00% accuracy, respectively, outperforming leading LLMs and specialized ToM methods. Additional experiments demonstrate improvements of 28.7, 39.3, and 16.6 percentage points in mental state and behavior prediction, long-sequence prediction, and implicit mental state inference, respectively. The results demonstrate that MMEA provides an interpretable framework for continuous mental modeling, enabling AI to move beyond surface-level semantic representations to infer individuals' underlying mental states from linguistic information and dynamically simulate their internal cognitive processes and evolution over time. MWM also provides a foundation for causal reasoning, personalized AI, sustained human–AI interaction, and future advances toward artificial general intelligence (AGI).

## 1 INTRODUCTION

Despite substantial advances in language understanding and reasoning enabled by LLMs, modeling human internal mental states and their dynamic evolution remains a fundamental challenge. Existing ToM approaches typically represent mental states through a limited set of variables, such as beliefs, goals, intentions, and emotions, but fail to adequately capture how subjective cognition emerges from prior experiences and subsequently influences mental states and behavior. Unlike the physical world, the human mental world is inherently subjective, largely implicit, and not directly observable. Consequently, identical external events may induce distinct psychological responses depending on an individual's experiential history, cognitive patterns, and existing mental states. Addressing these limitations requires moving beyond isolated mental state inference based primarily on linguistic and semantic representations. Inspired by the principles of world models, particularly explicit internal state representation, dynamic simulation, and state evolution, we argue for a computational framework that explicitly models and simulates the internal mental world, capturing the interactions among subjective cognition, mental states, and their temporal evolution to enable continuous and interpretable modeling of the end-to-end process underlying behavioral generation.

To address this challenge, we propose the MWM and its core implementation, the MMEA. Drawing on Eastern and Western psychological theories, MMEA explicitly constructs and continuously updates an individual's internal mental world through multiple branches of subjective interpretation, fine-grained mental state representations, cognitive biases, and mental inertia. This enables continuous and interpretable mental state modeling and behavior prediction. Across public benchmarks and three experiments on self-constructed datasets covering continuous mental state tracking, history-sensitive differentiation, implicit mental state inference, and ablation studies, MMEA demonstrates substantial advantages over existing LLMs and specialized LLM-based ToM reasoning frameworks.

# 2 RELATED WORK

## 2.1 Theory of Mind and Psychological Dynamics in Psychology and Cognitive Science

Western and Eastern psychology provide complementary foundations for understanding subjective cognition, mental states, and their evolution. Western psychology emphasizes cognitive representations and emotional processes, while Eastern psychology focuses on interacting mental factors and historical influences.

In Western psychology, mental model theory highlights the role of prior knowledge in subjective interpretation [13], while cognitive theories of emotion and the Component Process Model (CPM) explain emotional responses through subjective appraisal and interacting psychological processes [14].

In Eastern psychological traditions, self-related biases and interpretations of external objects shape perception, mental factors interact to produce psychological states [25], [26], and accumulated experiences influence subsequent cognition [25] – [27].

## 2.2 Mental Modeling, Cognitive Computing, and World Models in Artificial Intelligence

Mental modeling in AI has evolved from static state recognition toward cognitive reasoning and dynamic modeling. Cognitive computing, affective computing, and world models provide foundations for modeling internal psychological processes through cognitive mechanisms, mental state representations, and state evolution.

Cognitive computing integrates knowledge representations, reasoning mechanisms, cognitive biases, personality, and historical memory to model belief formation and behavior [15], [16]. Affective computing has advanced toward fine-grained emotion understanding and dynamic state modeling through multimodal representations and cognitive appraisal [28]– [30].

World models learn latent state representations and transition dynamics for simulation and prediction [17]. Recent approaches extend these mechanisms to social cognition by modeling social states, agent behavior, and belief changes [8], [9]. Together, these developments provide a foundation for computational models of evolving internal psychological processes.

## 2.3 Limitations of Existing Research

Despite advances in ToM, cognitive computing, affective computing, and world models, existing approaches still rely largely on linguistic and behavioral cues to infer mental states, with limited systematic modeling of how individuals' internal mental worlds form and evolve. Three main limitations remain.

First, existing methods lack explicit representations of subjective cognition, often overlooking how prior beliefs, cognitive biases, and personal experiences shape individual interpretations of events. Second, cognition and mental states are frequently modeled separately, with limited consideration of psychological interactions, temporal continuity, and mental inertia, making it difficult to capture complex mental processes and their evolution. Third, memory mechanisms primarily focus on historical information retrieval rather than modeling how past experiences establish persistent cognitive biases and influence subsequent mental states through long-term cognitive changes and mental inertia.

These limitations constrain the simulation of internal psychological processes, long-term mental evolution, and behavior prediction under historical dependencies. A unified computational framework is therefore needed to integrate subjective world construction, cognitive–psychological interactions, dynamic state evolution, and historical dependencies, enabling continuous modeling and simulation of internal mental processes.

# 3 METHODS

## 3.1 Theoretical Foundations of the Mental World Model

Eastern and Western psychology and cognitive science provide theoretical foundations for mental modeling through cognitive mechanisms, psychological structures, and dynamic evolution. However, conventional world models primarily learn environmental dynamics from external observations, making them difficult to apply directly to the subjective, implicit, and unobservable human mental world. To address this limitation, we propose the MWM and its reasoning architecture, the MMEA, which integrates subjective cognition, mental states, and their transition dynamics into a unified computational framework. MMEA explicitly represents subjective cognition and mental states, simulates interactions among psychological factors and their continuous evolution shaped by historical experiences, and predicts emotional and behavioral responses.

## 3.2 Architecture of the Mental World Model

### 3.2.1 Overall Framework

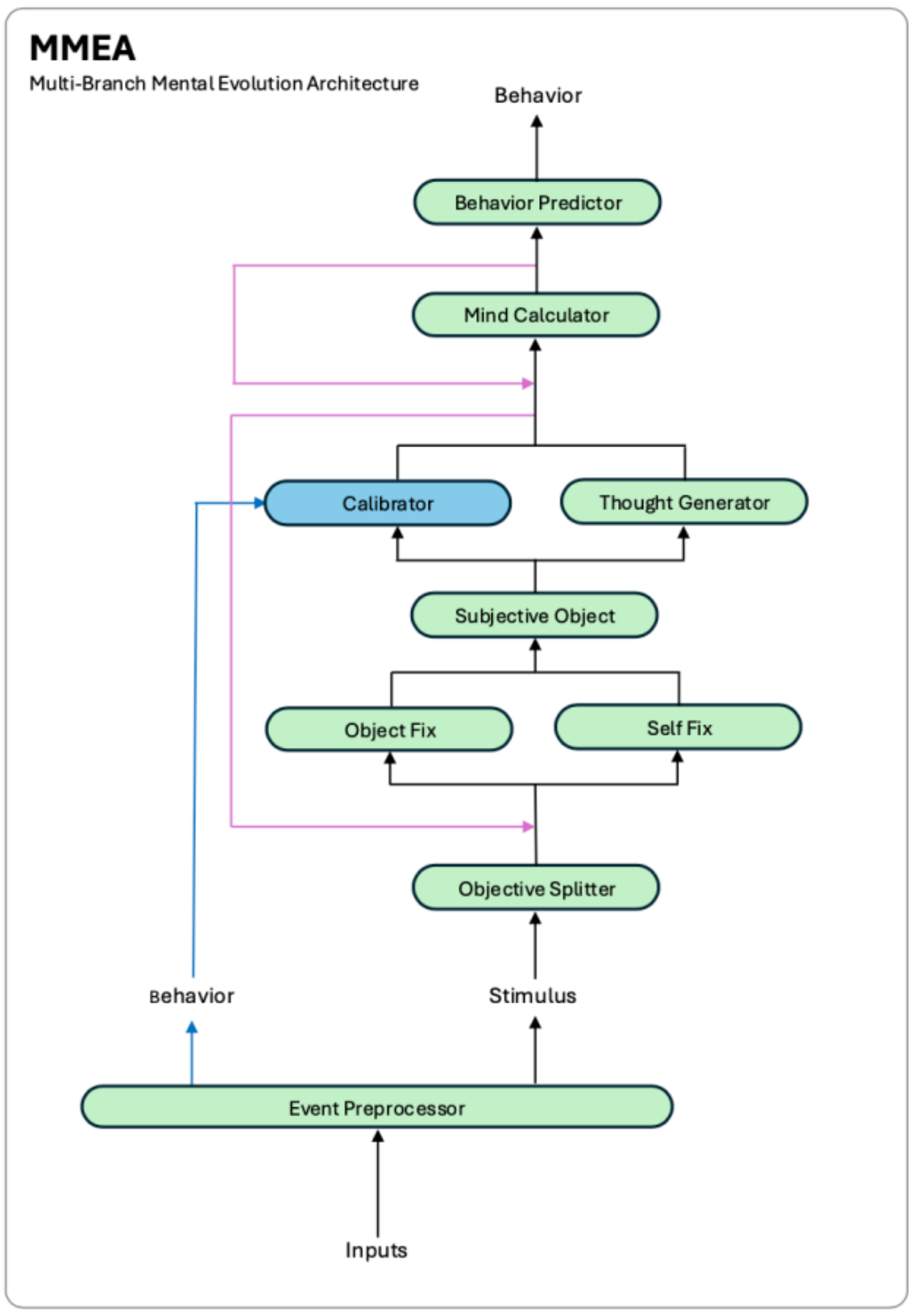

Figure 1: Architecture of MMEA

The MWM aims to provide a unified representation of an individual's subjective cognition, mental states, and behavioral tendencies across a sequence of events. To this end, we develop the MMEA, which integrates subjective world construction, cognitive appraisal, mental factor computation, historical continuity modeling, and behavior prediction into a unified dynamic computational framework.

The model takes a sequence of events and a specified target individual as input. The event sequence is defined as:

$$\mathcal{E} = \{E_1, E_2, \dots, E_T\} \tag{1}$$

Where $E_T$ denotes the event at time step t, and T is the total number of events. The first T-1 events are used to interpret the individual's observed actions and update historical states, while the final event $E_T$ is used to predict behavior that has not yet occurred.

The core computational process of MMEA follows a hierarchical information flow:

$$E_t \xrightarrow{\text{Grasping}} \mathcal{C}_t \xrightarrow{\text{Perception}} \mathcal{O}_t \xrightarrow{\text{Cognition}} \mathcal{T}_t \xrightarrow{\text{Mental Calculation}} \mathcal{S}_t \xrightarrow{\text{Prediction}} P(A_t) \tag{2}$$

Where $\mathcal{C}_t$ denotes the set of candidate Self Fix and Object Fix combinations, $\mathcal{O}_t$ represents the set of subjective world branches with different prior activation probabilities, $\mathcal{T}_t$ denotes the Thoughts (cognitive appraisals) associated with each branch, $\mathcal{S}_t$ represents the set of 40-dimensional mental factor states across branches, and $P(A_t)$ denotes the final behavioral probability distribution. MMEA performs cognitive and psychological computations independently within each subjective world branch and integrates their outputs according to branch probabilities to obtain the final prediction.

To maintain continuity across events, MMEA incorporates two distinct mechanisms: Cognitive Bias $H$ and Mental Inertia $M$. Cognitive Bias $H$ incorporates historical Self Fix and Object Fix activations into the construction of the current subjective world, further influencing Thought generation within the corresponding branches. Mental Inertia $M$, in contrast, integrates historical mental states into each branch after the current mental factor computation. These mechanisms capture two distinct forms of historical influence: how past experiences shape the subjective interpretation of current events, and how previous mental states continue to affect present psychological responses. Their relationship can be expressed as:

$$
\begin{aligned}
\mathcal{C}_t &= f_{\text{grasp}}(E_t, H_{t-1}),\\
\mathcal{O}_t &= f_{\text{world}}(\mathcal{C}_t, E_t),\\
\mathcal{T}_t &= f_{\text{thought}}(\mathcal{O}_t, H_{t-1}),\\
\mathcal{S}_t &= f_{\text{mental}}(\mathcal{O}_t, \mathcal{T}_t),\\
\tilde{\mathcal{S}}_t &= f_{\text{inertia}}(\mathcal{S}_t, M_{t-1}),\\
P(A_t) &= f_{\text{behavior}}(\mathcal{O}_t, \mathcal{T}_t, \tilde{\mathcal{S}}_t).
\end{aligned} \tag{3}
$$

During behavior prediction, MMEA combines branch-specific behavioral distributions weighted by their prior activation probabilities.

Through this architecture, MMEA integrates subjective world construction, multi-branch cognitive and psychological computation, historical continuity, and behavior generation into a unified process capable of dynamic evolution. It also preserves intermediate representations, including Self Fix and Object Fix, subjective interpretations, cognitive appraisals, mental factors, and behavioral probabilities, providing a traceable computational pathway for mental reasoning and behavior prediction.

### 3.2.2 Hierarchical Perception and Subjective World Construction

The primary objective of the perception layer is to transform objective external events into an individual's subjective world. Since the same event may carry different meanings for different individuals, the model requires an explicit mechanism for translating objective stimuli into subjective interpretations. To achieve this, MMEA employs a hierarchical perception structure consisting of five levels: L0, L1, L2, L2.5, and L3, responsible for event decomposition, fact extraction, cognitive fixation, combination activation, and subjective world generation, respectively.

At L0, events are decomposed into external stimuli and observed actions. At L1, observable facts are extracted to constrain subjective interpretations. At L2, the model constructs Self Fix, representing cognitive attachments to people, objects, and relational positions, and Object Fix, representing stable interpretations of events and rules. Cognitive candidates are generated through Fresh (current events) and Historical (prior biases) pathways. After merging and deduplication, Fresh candidates take precedence to prevent historical biases from overriding new interpretations.

At L2.5, the model combines Self Fix and Object Fix candidates and computes their activation strengths. For a given combination $C_j$, let $I_j$ denote the confidence of its Self Fix, $D_j$ the confidence of its Object Fix, and $g_j$ its compatibility with the current event. The activation strength is then calculated as follows:

$$
q(C_j) = I_j D_j g_j \tag{4}
$$

$$
P(C_j) = \frac{q(C_j)}{\sum_{C_k \in \mathcal{C}_t^*} q(C_k)} \tag{5}
$$

Where $\mathcal{C}_t^*$ denotes the set of valid combinations. This mechanism prevents low-compatibility combinations from being assigned a probability of 1 simply because no competing candidates are available. After that, the model typically retains at most three cognitive branches to balance interpretive diversity and computational cost.

$$
P(O_j) = P(C_j) \tag{6}
$$

At L3, each valid combination is transformed into a Subjective Object $O_j$, with its prior probability inherited from the corresponding combination's activation probability.

### 3.2.3 Multi-Branch Cognitive Reasoning and Mental State Computation

MMEA performs cognitive reasoning and mental state computation independently within each subjective world branch before aggregating the results across branches. Each branch maintains its own cognitive interpretation and 40-dimensional mental state representation, preserving distinct psychological pathways and preventing premature information loss.

For mental state computation, MMEA decomposes mental states into 40 distinct mental factors, each evaluated independently within every branch. The aggregate score for mental factor $f$ is defined as:

$$
s_b(f) = 0.35 s_{1,b}(f) + 0.30 s_{2,b}(f) + 0.10 s_{3,b}(f) + 0.25 s_{4,b}(f) \tag{7}
$$

The four dimensions correspond to Subjective Object, Thought, Volition, and a supplementary assessment based on modern psychology. Subjective Object and Thought receive the highest weights, reflecting the central role of subjective cognition in the emergence of mental factors. Volition is assigned a lower weight to prevent observed actions from disproportionately influencing mental state inference, while the modern psychological assessment provides complementary information.

The event-level mental state is obtained by aggregating the mental factor scores across branches using their respective probabilities:

$$M_{\text{event}}(f) = \sum_{b=1}^{B} a_b \, s_b(f)$$
$$a_b = \frac{P(O_b)}{\sum_{k=1}^{B} P\,(O_k)} \quad (8)$$

The action-adjusted posterior is used to update historical Cognitive Bias, while the prior probabilities preserve the relative activation strengths of different subjective interpretations of the current event. This distinction separates cognitive activation from corrections based on behavioral evidence.

### 3.2.4 Cross-Event Cognitive Bias and Mental Inertia Modeling

An individual's psychological processes are continuously influenced by existing cognitive tendencies and historical mental states. To capture these effects, MMEA employs two independent mechanisms: Cognitive Bias $H$ and Mental Inertia $M$. The former models how past experiences shape subjective interpretations, while the latter captures the residual influence of previous mental states on current psychological responses.

Cognitive Bias $H$ stores Subjective Objects, Self Fix, Object Fix, and their associated activation strengths $m$ formed through previous events. After each event, the model records historical cognitive biases using the action-adjusted posterior probabilities. When a new bias is semantically similar to an existing record, its strength is reinforced according to the following rule:

$$m_{\text{new}} = m_{\text{old}} + \alpha(1 - m_{\text{old}})m_{\text{incoming}} \quad (9)$$

Where $\alpha$ = 0.35. This update rule progressively strengthens recurring cognitive tendencies while limiting the magnitude of reinforcement to prevent the strength from exceeding its upper bound.

For each new event, up to five relevant historical biases are retrieved and incorporated into the L1 and L2 stages to influence subjective world construction.

Mental Inertia $M$ preserves historical mental states and their residual effects on subsequent events. To prevent repeated accumulation of historical influences, the model stores the original $M_{\text{event}}$ generated by each event rather than the final state after historical integration.

Let $\theta_i$ denote the importance of historical event $i$, and $d_i$ its temporal distance. The decay timescale and historical weight are defined as:

$$\tau(\theta_i) = 3 + 57\theta_i$$
$$w_i = \theta_i \exp\left(-\frac{d_i}{\tau(\theta_i)}\right) \quad (10)$$

This formulation assigns more important events a longer period of psychological influence, with decay timescales ranging from 3 to 60 days. Temporal decay is applied only when the time interval between events is explicitly given or can be reasonably inferred; otherwise, $d_i$=0.

The historical mental state is computed as a weighted average:

$$M_{\text{hist}}(f) = \frac{\sum_i w_i \, M_{\text{event}}^{(i)}(f)}{\sum_i w_i} \quad (11)$$

The final mental state is then given by:

$$M_{\text{final}}(f) = (1 - \lambda)M_{\text{event}}(f) + \lambda M_{\text{hist}}(f) \quad (12)$$

Here, $\lambda$ controls the influence of historical Mental Inertia and is capped at 0.5, preventing previous mental states from completely overriding the psychological response generated by the current event.

### 3.2.5 Multi-Branch Behavior Generation and Probabilistic Prediction

Behavior prediction is the final stage of MMEA, in which internal mental states are mapped to externally observable actions. Rather than generating behavior directly from event descriptions, the model independently computes mental states and behavioral tendencies within each subjective world branch before aggregating their predictions across branches.

Within each branch, the model generates observable behavioral candidates based on the Subjective Object, Thought, and mental state after historical integration. Behavior generation proceeds in two stages. First, the model generates concrete, distinguishable actions and removes semantic duplicates. It then estimates the conditional probability of each candidate behavior:

$$P\left(A_j \mid O_b, T_b, \tilde{s}_b\right) \quad (13)$$

The final behavioral probability distribution is obtained by weighting the branch-specific predictions according to their prior activation probabilities:

$$P(A_j) = \sum_{b=1}^{B} a_b \, P(A_j \mid O_b, T_b, \tilde{s}_b) \qquad (14)$$

Where $a_b$ denotes the prior activation probability of branch. The model then merges semantically equivalent behaviors across branches, retains the five candidates with the highest probabilities, and produces a normalized behavioral probability distribution.

Overall, MMEA integrates individual subjective cognition, psychological changes, and behavior generation into a mental simulation framework through hierarchical subjective world construction, multi-branch cognitive and psychological computation, two complementary mechanisms of historical continuity, and prediction feedback. The resulting framework supports history-dependent, dynamically continuous, and traceable mental modeling.

# 4 EXPERIMENTS AND RESULTS

## 4.1 Experimental Design and General Setup

This study uses public benchmarks, custom experiments, and ablation studies to systematically evaluate MMEA's capabilities in dynamic mental reasoning, continuous mental state tracking, historical dependency modeling, and behavior prediction.

This study evaluates MMEA on two public benchmarks, DynToM and RecToM, comprising 1,820 questions. DynToM assesses changes in beliefs, emotions, intentions, and behaviors through four subtasks: How, What, Whether, and Why. RecToM evaluates behavior prediction from multi-turn interactions between Recommenders and Seekers. MMEA is compared with four general-purpose LLMs (Gemini-3.8-Flash, GPT-6 Luna, Grok 4.3, and DeepSeek-V4-Flash) and five cognitive reasoning frameworks (CoT, SimToM, TimeToM-style, PICTURE, and Agentic-ToM) to assess dynamic mental reasoning and behavior prediction.

Beyond the public benchmarks, three additional experimental datasets are constructed to investigate whether MMEA can move beyond surface-level semantic representations to model, simulate, and track the evolution of individuals' internal mental worlds. The first dataset uses continuous event sequences involving psychological conflicts, mixed emotions, and implicit motivations to evaluate mental state tracking and behavior prediction. The second dataset uses paired event sequences with subtle historical differences to assess the model's ability to capture historical dependencies and divergent psychological trajectories. The third dataset examines scenarios in which observable behavior conflicts with internal mental states, evaluating whether the model can infer underlying mental states beyond surface-level semantic cues. Finally, ablation experiments are conducted to quantify the contribution of individual components and evaluate the effectiveness and necessity of the proposed architecture.

## 4.2 Experiment 1: Dynamic Mental Reasoning on DynToM

To evaluate MMEA's ability to model dynamic mental states and perform continuous mental reasoning, we conduct experiments on DynToM, a public benchmark for dynamic Theory of Mind. We evaluate 1,420 questions from 20 test sets, each containing 71 questions, across four subtasks: How, What, Whether, and Why.

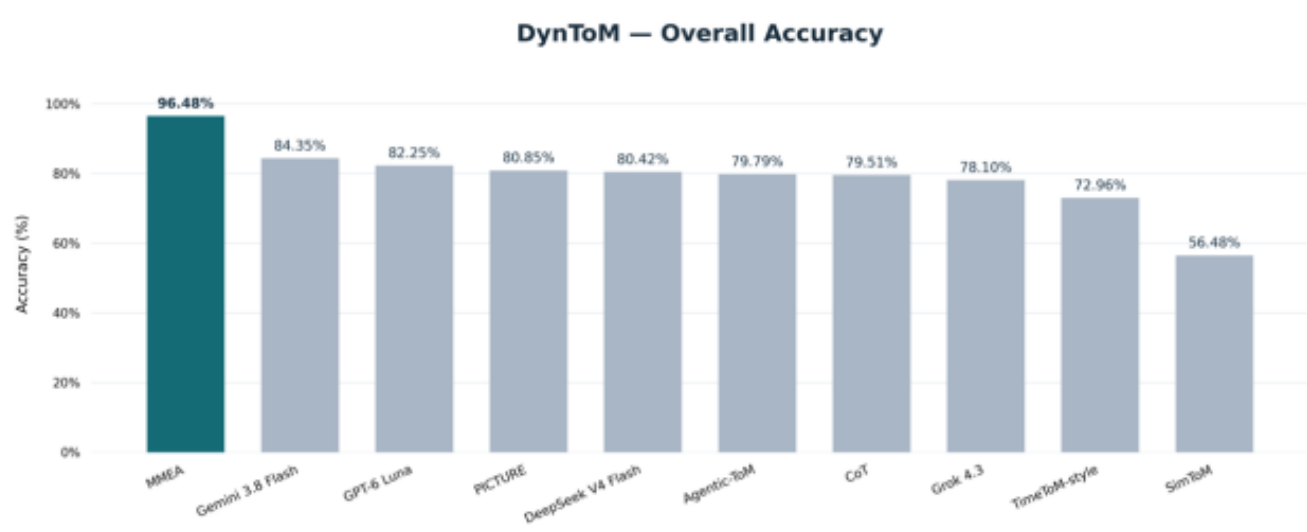

Figure 2: DynToM – Overall Accuracy

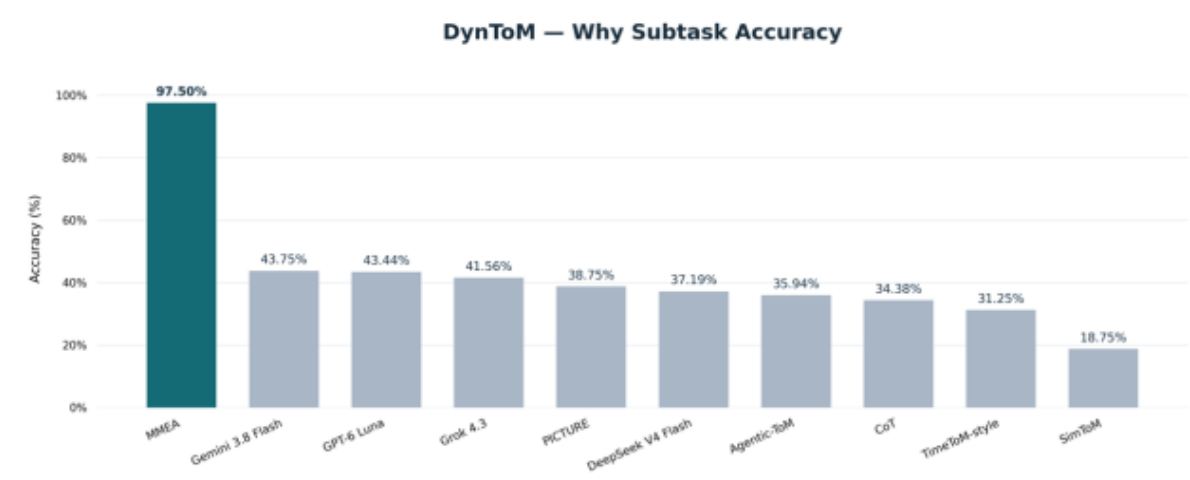

Figure 3: DynToM – Why Subtask Accuracy

As shown in Figure 2, MMEA achieves an overall accuracy of 96.48% on the 1,420 questions in the DynToM dynamic mental reasoning benchmark, substantially outperforming the best-performing general-purpose LLM,

Gemini-3.8-Flash (84.35%), and the best-performing specialized ToM reasoning method, PICTURE (80.85%). These results demonstrate MMEA's overall advantage in dynamic mental reasoning. Conventional LLMs and existing ToM methods primarily infer mental states from linguistic information and often struggle to maintain consistent mental representations across events, even when incorporating chain-of-thought prompting, perspective-taking, or multi-stage reasoning. In contrast, MMEA explicitly models subjective cognition, mental states, and historical dependencies, representing psychological changes as a continuous and causally connected process. This enables more stable and accurate dynamic mental reasoning.

As shown in Figure 3, MMEA achieves a more pronounced advantage on the Why subtask, which evaluates the ability to explain changes in mental states. MMEA reaches an accuracy of 97.50%, compared with 43.75% for the best-performing general-purpose LLM, Gemini-3.8-Flash, and 35.94% for the best-performing specialized ToM method, Agentic-ToM. The Why subtask requires models to explain how mental states arise and how previous experiences influence their development, placing greater demands on causal reasoning across events. The substantial performance decline of existing models on this subtask highlights their limitations in connecting past experiences, subjective interpretations, and psychological changes. By preserving subjective cognition, historical cognitive biases, and evolving mental states, MMEA supports reasoning based on continuous psychological trajectories, demonstrating its advantage in modeling the formation of mental states and their evolving causal relationships.

## 4.3 Experiment 2: Interactive Behavior Prediction on RecToM

Following the evaluation of dynamic mental reasoning, we further assess MMEA's ability to predict interaction strategies and subsequent behavior from internal mental representations using the public RecToM benchmark. RecToM consists of multi-turn dialogues in recommendation scenarios, requiring models to predict participants' subsequent behavior based on dialogue history, individual preferences, and current intentions. We evaluate 400 questions across two tasks, Recommender and Seeker, with 200 questions each, assessing the prediction of subsequent recommendation strategies and user responses, respectively. Consistent with Experiment 1, we compare MMEA against general-purpose LLMs and specialized ToM reasoning methods to evaluate the effectiveness of explicit mental modeling in behavior prediction tasks relevant to real-world interactions.

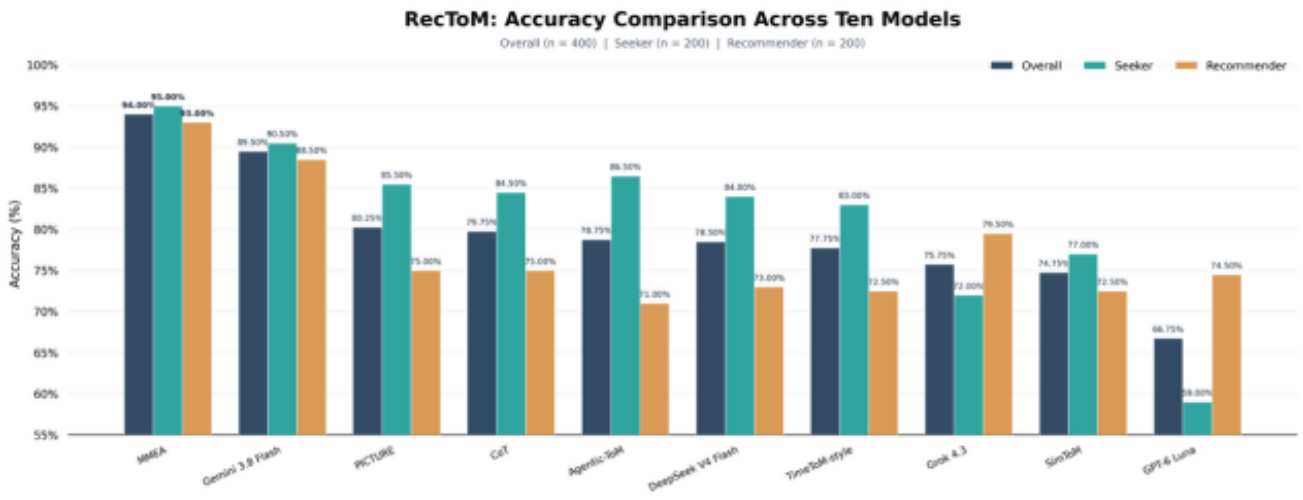

Figure 4: RecToM Performance Comparison Across Overall and Subtask Accuracy

As shown in Figure 4, MMEA achieves an overall accuracy of 94.00% on the 400 questions in the RecToM benchmark, outperforming the best-performing general-purpose LLM, Gemini-3.8-Flash (89.50%), and the best-performing specialized ToM reasoning method, PICTURE (80.25%). This result indicates that MMEA can effectively translate internal mental representations into predictions of subsequent behavior. While conventional LLMs primarily rely on linguistic cues and interaction patterns in dialogue, MMEA continuously maintains subjective cognition, historical dependencies, and mental states, allowing behavioral predictions to incorporate the individual's evolving internal psychological processes.

Across the two interaction tasks, MMEA achieves accuracies of 93.00% for Recommender and 95.00% for Seeker, showing consistent predictive performance across roles. Recommenders must select response strategies based on the other participant's needs and the progression of the interaction, whereas Seekers determine their subsequent actions based on personal preferences and previous feedback. MMEA's performance on both tasks suggests that its internal mental representations can accommodate different role-specific decision processes and support the prediction of behavioral tendencies in complex interactive contexts.

## 4.4 Experiment 3: Continuous Mental State Tracking and Behavior Prediction

This experiment evaluates the ability of models to dynamically represent and track an individual's cognition and mental states across successive events and to predict subsequent behavior based on accumulated history. The experiment consists of a sequence of temporally connected events. At each event, the model first predicts the target individual's Thoughts and Actions. The action assigned the highest predicted probability is then treated as the action that occurred, incorporated into the event history as a new objective fact, and used as context for the next prediction. Through this iterative procedure, the individual's experiences, cognitive interpretations, mental states, and behaviors progressively accumulate and evolve.

We evaluate MMEA, DeepSeek-V4-Flash, Grok 4.3, GPT-6 Luna, and Gemini-3.8-Flash using identical event sequences and compare their mental state and behavior predictions. Four evaluation metrics are used to assess overall prediction accuracy, the stability of consecutive correct predictions, the ability to predict complete trajectories without errors, and predictive performance at different stages of a sequence.

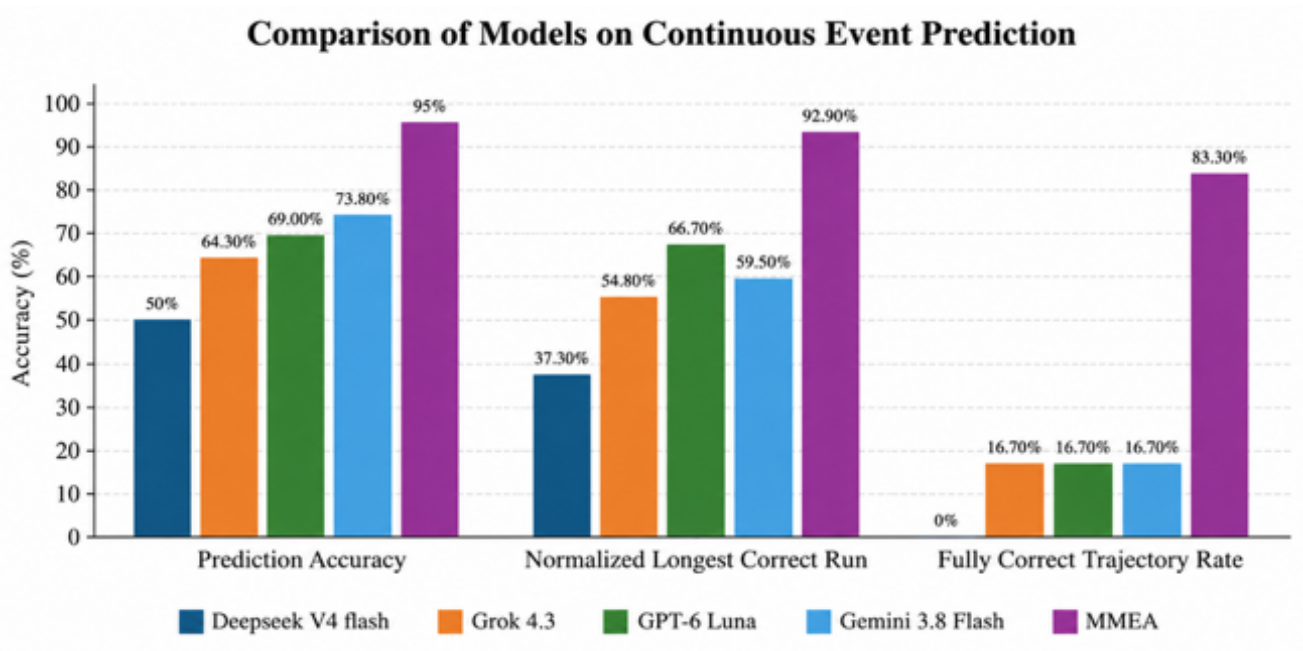

Figure 5: Comparison of Models on Continuous Event Prediction

Prediction Accuracy (PA) measures the overall accuracy of mental state and behavior prediction as the proportion of correct predictions across all test events, with higher values indicating better performance, and is defined as follows:

$$\mathrm{PA} = \frac{1}{N}\sum_{i=1}^{N} \mathbb{I}\left(\hat{y}_i = y_i\right) \qquad (15)$$

Where $N$ denotes the total number of test events, $\hat{y}_i$ is the model's prediction for event $i$, $y_i$ is the corresponding ground-truth answer, and $\mathbb{I}(\cdot)$ is an indicator function that equals 1 when the prediction is correct and 0 otherwise.

Experimental results show that MMEA achieves a PA of 95.0%, substantially outperforming Gemini-3.8-Flash (73.8%). These results indicate that MMEA provides more accurate predictions across the evaluated events, demonstrating a stronger ability to infer mental states and subsequent behaviors from an individual's history and current circumstances.

Normalized Longest Correct Run (NLCR) measures the stability of mental state tracking and behavior prediction across consecutive events by calculating the normalized length of the longest sequence of correct predictions within each trajectory, and is defined as follows:

$$\mathrm{NLCR} = \frac{1}{M}\sum_{j=1}^{M} \frac{L_j^{\max}}{T_j} \qquad (16)$$

Here, $M$ is the total number of test trajectories, $T_j$ is the number of events in trajectory $j$, and $L_j^{\max}$ is the longest sequence of consecutive correct predictions.

MMEA achieves an NLCR of 92.9%, compared with 66.7% for GPT-6 Luna. These results show that MMEA maintains longer uninterrupted sequences of correct predictions, whereas the comparison models exhibit more frequent prediction errors as events progress. This suggests that MMEA provides greater stability in long-term mental state tracking and continuous behavior prediction.

Fully Correct Trajectory Rate (FCTR) measures long-term predictive reliability as the proportion of trajectories in which all events are correctly predicted, assigning a score of 1 if every prediction is correct and 0 otherwise:

$$\mathrm{FCTR} = \frac{1}{M}\sum_{j=1}^{M} \mathbb{I}\left(\prod_{t=1}^{T_j} c_{j,t} = 1\right) \qquad (17)$$

Here, $M$ is the total number of test trajectories, $T_j$ is the number of events in trajectory $j$, and $c_{j,t} \in \{0,1\}$ indicates whether the prediction for event $t$ is correct.

MMEA achieves an FCTR of 83.3%, compared with 16.7% for Gemini-3.8-Flash. These results show that MMEA maintains correct predictions throughout most complete trajectories, whereas the comparison models typically make at least one error within a trajectory. This finding further suggests that MMEA is more effective at integrating historical information and maintaining consistent predictions across entire event sequences.

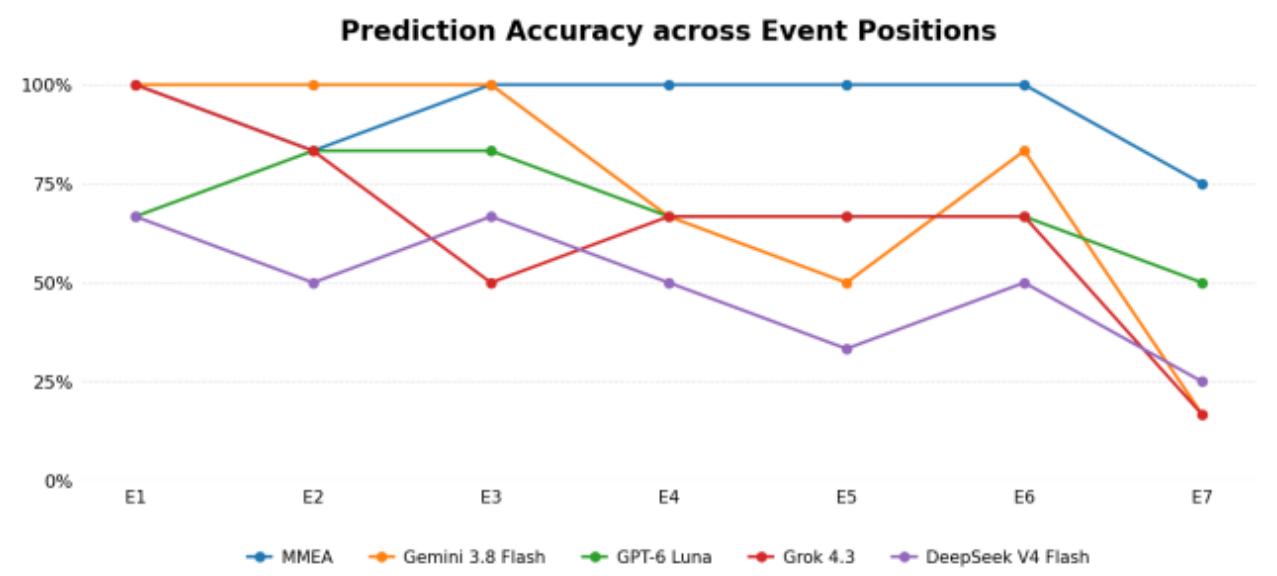

Figure 6: Prediction Accuracy across Event Positions

Event-wise Accuracy (EA) measures the model's ability to use historical information as individual experiences, interpersonal relationships, and mental states accumulate over time by calculating prediction accuracy at each position in a continuous event sequence, and is defined as follows:

$$EA_t = \frac{\sum_{k \in \mathcal{K}_t} c_{k,t}}{|\mathcal{K}_t|} \tag{18}$$

Where $\mathcal{K}_t$ denotes the set of valid trajectories containing event position $t$.

Experimental results show that MMEA consistently achieves higher prediction accuracy than the other LLMs across event positions and maintains relatively stable performance as sequences become longer. In contrast, the comparison models generally exhibit a noticeable decline in accuracy over longer sequences. These findings suggest that MMEA provides greater stability in continuous mental state tracking and long-term behavior prediction.

## 4.5 Experiment 4: History-Sensitive Differentiation of Mental States and Behavior

This experiment evaluates whether models can detect subtle historical differences and preserve their influence on subsequent mental state and behavior predictions. We construct two nearly identical event trajectories, A and B, differing only in a critical historical detail that affects subjective interpretation, while keeping all subsequent events identical to isolate the effect of history.

We use the Psychological Trajectory Score (PTS) to map model outputs onto a directional scale ranging from −2 to +2. Positive and negative values indicate the psychological trajectories associated with histories A and B, respectively, while their absolute magnitudes reflect the strength of the corresponding tendency. A score of 0 indicates no clear directional tendency. Human-annotated reference trajectories, denoted as True A and True B, are included for comparison.

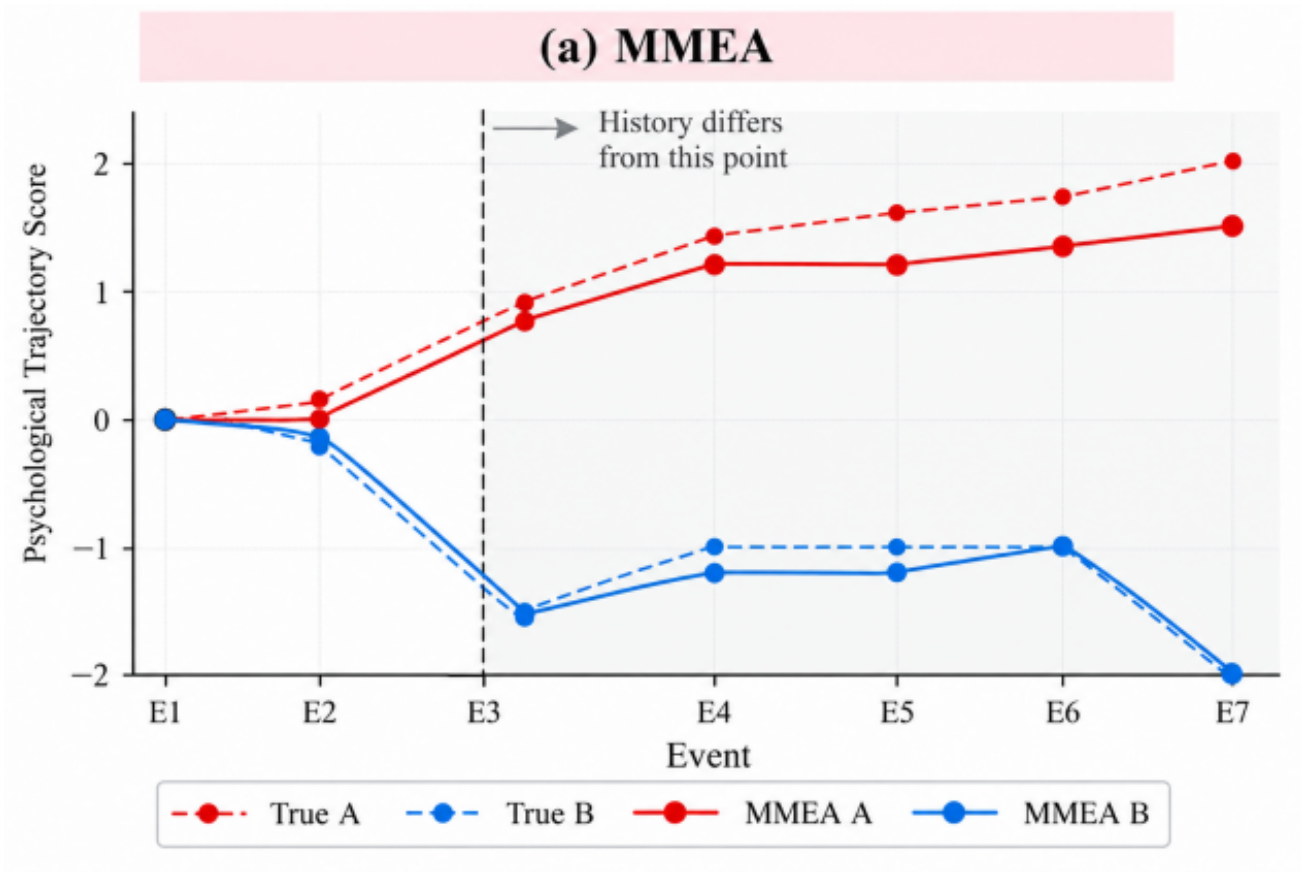

Figure 7: Psychological Trajectory Differentiation by MMEA

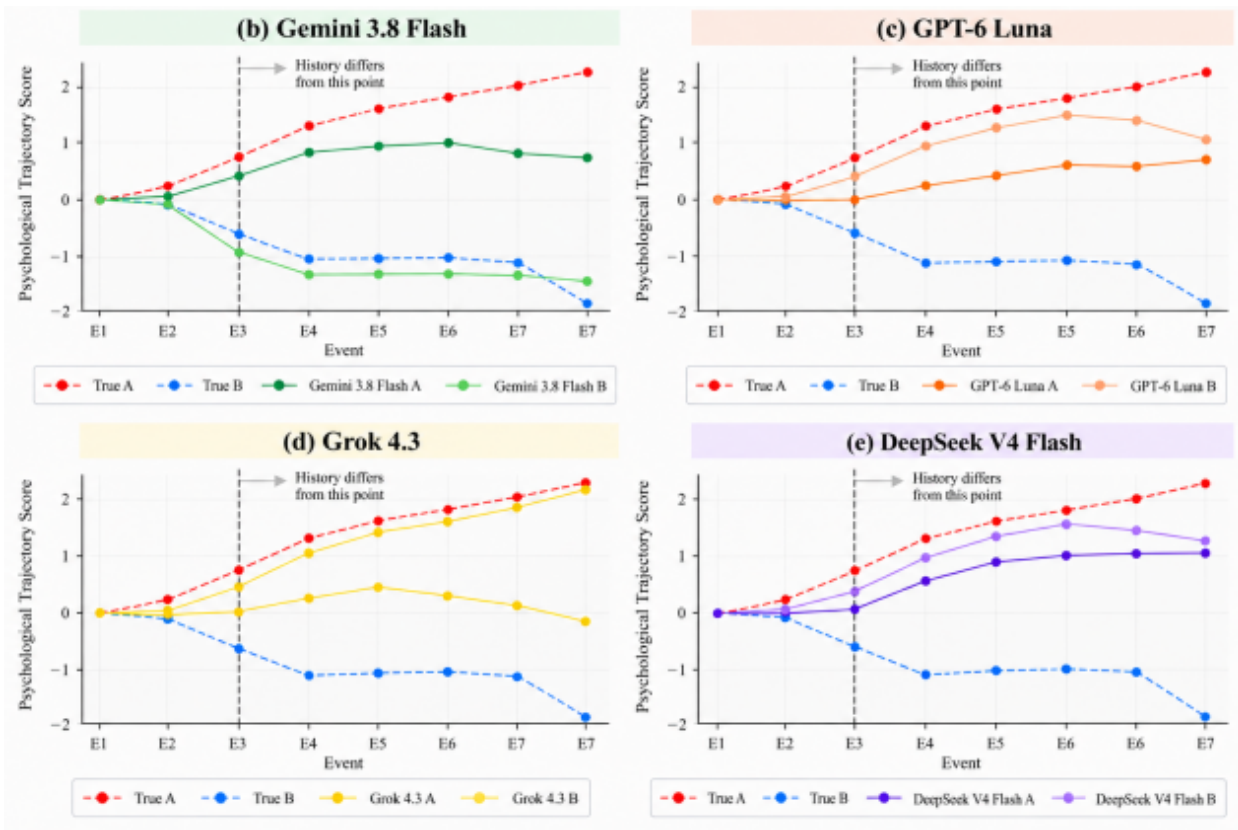

Figure 8: Predicted Psychological Trajectory Differentiation by Four Other Large Language Models

As shown in Figure 7, the two reference trajectories remain largely consistent during E1–E2. Following the critical historical difference, they begin to diverge substantially, reaching 2.0 and −2.0 at E7. MMEA closely tracks this divergence, with its A/B trajectories progressing from 0.8−1.6 at E3 to 1.5−2.0 at E7, remaining broadly consistent with the reference trajectories. Gemini-3.8-Flash also preserves the correct bidirectional divergence, although with a smaller magnitude. In contrast, the B trajectories predicted by GPT-6 Luna, Grok 4.3, and DeepSeek-V4-Flash gradually shift toward the same positive direction as trajectory A after the historical difference is introduced, failing to preserve the opposing psychological trajectories associated with the two histories.

These results suggest that MMEA can retain subtle historical differences within continuously evolving mental states, allowing them to influence the interpretation and prediction of subsequent identical events. In contrast, several general-purpose LLMs tend to produce converging psychological trajectories across different historical conditions as the event sequence progresses.

### 4.6 Experiment 5: Implicit Mental State Inference

This experiment treats each event as an independent test case, focusing on scenarios involving concealed emotions, discrepancies between expressed and actual intentions, and inconsistencies between words and actions. It evaluates whether models can infer underlying cognitive and mental states beyond surface-level linguistic and behavioral cues.

To assess both the plausibility of inferred Thoughts and their predicted probabilities, we introduce the Probability-Weighted Thought Score (PWTS), where higher scores indicate that the model assigns greater probability to inferences that more closely match the individual's underlying mental state, as defined below:

$$PWTS = \frac{1}{N}\sum_{i=1}^{N}\sum_{j=1}^{K_i} p_{ij}\, s_{ij} \qquad (19)$$

Where $N$ denotes the total number of test events, $K_i$ is the number of Thoughts generated for event $i$, $p_{ij}$ is the probability assigned by the model to the $j$-th Thought, and $s_{ij}$ represents the degree of agreement between that Thought and the reference psychological inference.

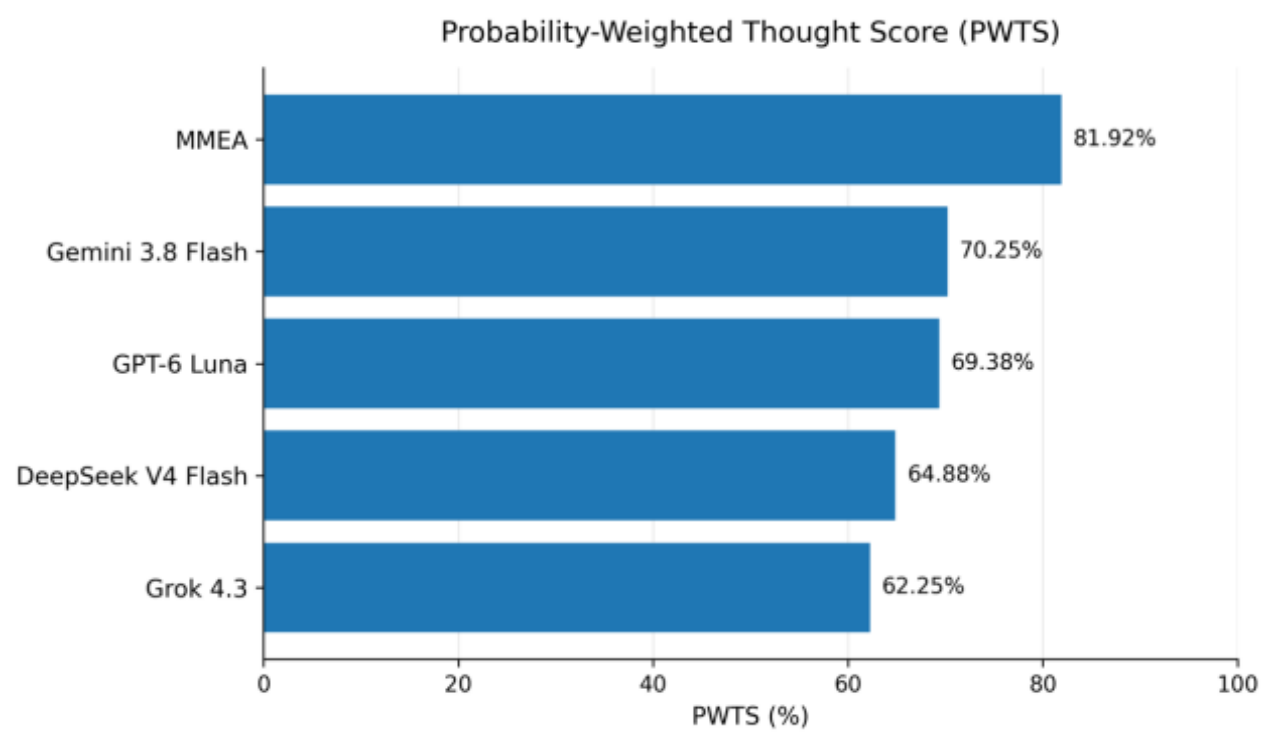

Figure 9: PWTS Results for Five Models

Figure 9 shows that MMEA achieves an average PWTS of 81.92%, outperforming the best-performing general-purpose LLM, Gemini-3.8-Flash (70.25%), by 11.67 percentage points. These results suggest that MMEA can move beyond surface-level linguistic and behavioral cues to infer underlying mental states. In situations involving concealed emotions, conflicting words and actions, or unexpressed intentions, MMEA demonstrates a stronger ability to model internal psychological processes rather than relying solely on semantic patterns.

### 4.7 Experiment 6: Ablation Study of Key Architectural Components

We conduct an ablation study to assess the contributions of MMEA's key components to its overall mental modeling performance. Using identical test data and evaluation procedures, we separately remove the Object Fix and Self Fix, Mind Calculator, and Subjective Object modules. Performance is evaluated using three metrics introduced in the continuous prediction experiment: PA, NLCR, and FCTR.

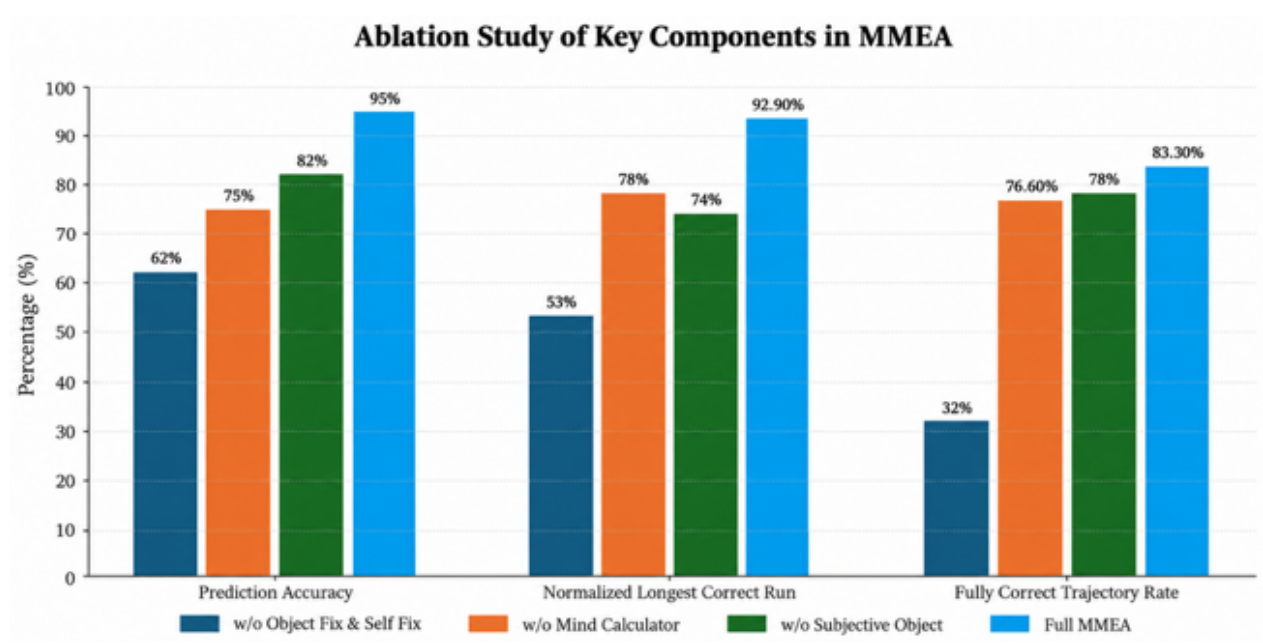

Figure 10: Ablation Study of Key Components in MMEA

The full MMEA architecture achieves PA, NLCR, and FCTR scores of 95.0%, 92.9%, and 83.3%, respectively, outperforming all ablated variants. Removing Object Fix and Self Fix results in the largest performance decline, with the three metrics dropping to 62.0%, 53.0%, and 32.3%. This indicates that continuously updating cognitive representations of external objects and the self is critical for long-term mental state evolution and predictive stability.

Overall, the ablation results indicate that the three components serve complementary roles in dynamic cognitive updating, mental state computation, and the continuity of subjective representations.

## 5 DISCUSSION

This section discusses its contributions, limitations, and future directions.

Results on DynToM and RecToM show that MMEA outperforms existing baselines in dynamic mental reasoning, explaining mental state changes, and behavior prediction. Experiment 3 demonstrates its ability to capture historical dependencies, maintain mental state continuity, and predict behavior over extended event sequences. Experiment 4 shows that MMEA can identify critical differences in long event sequences and capture their lasting effects on subsequent mental states and behavior. Experiment 5 further demonstrates MMEA's ability to move beyond semantic representations and simulate individuals' internal mental worlds. The ablation study in Experiment 6 indicates that dynamic cognitive and mental state updating, together with subjective world modeling, play complementary roles in long-term mental evolution and behavior prediction. Overall, these findings provide evidence for the feasibility and effectiveness of world model-based mental modeling. By dynamically representing and simulating internal mental processes, MMEA offers an alternative computational approach to mental modeling beyond conventional linguistic pattern matching.

However, several limitations remain. The current implementation of MMEA relies on LLMs to construct subjective worlds, compute mental states, and simulate their evolution. Its main contribution therefore lies in its mental modeling mechanisms and architectural design. Future work will explore learnable neural representations of Self Fix, Object Fix, Cognitive Fixation, and mental states, combined with internal state transition mechanisms to reduce reliance on LLM-based reasoning, computational costs, and inference latency. These developments may enable more autonomous and continuous mental modeling, with potential applications in personalized AI, affective computing, human–AI collaboration, and embodied intelligence, contributing to advances in AI social cognition and AGI research.

## 6 CONCLUSION

Drawing on Eastern and Western psychology and cognitive science, this paper introduces the MMEA, which explicitly constructs, maintains, and continuously updates an individual's subjective cognition and mental states. By integrating past experiences, cognitive interpretations, psychological changes, and behavior prediction, MMEA provides a unified framework for dynamic mental modeling. Its central contribution is to extend mental reasoning beyond direct inference from linguistic context toward the continuous representation and simulation of an individual's internal mental world, allowing the formation of subjective cognition, the evolution of mental states, and the generation of behavior to be modeled within a unified computational framework. Rather than focusing solely on predicting mental states and behavioral outcomes, MMEA preserves interpretable intermediate representations and traceable psychological trajectories, providing a computational approach to understanding historical dependencies, state continuity, and cognitive differences in complex psychological processes. Overall, this study establishes an initial computational and empirical foundation for mental world models capable of continuously simulating internal psychological processes beyond linguistic pattern matching and suggests a research direction for advancing AI from external behavioral inference toward deeper modeling of internal cognitive and psychological processes.

## REFERENCES

[1] Liu, B., Xu, D., Jiang, L., Guo, W., Chen, P. (2026). Modeling the mental world for embodied AI: A comprehensive review. arXiv preprint arXiv:2601.02378.

[2] Wang, L., Shelim, R., Saad, W., Ramakrishna, N. (2026). MetaMind: General and cognitive world models in multi-agent systems by meta-theory of mind. arXiv preprint arXiv:2603.00808.

[3] Tsvilodub, P., Klumpp, J.-F., Mohammadpour, A., Hu, J., Franke, M. (2026). On emergent social world models—Evidence for functional integration of theory of mind and pragmatic reasoning in language models. In ACL 2026, 64th Annual Meeting of the Association for Computational Linguistics, pages 37382–37420. Association for Computational Linguistics. doi:10.18653/v1/2026.acl-long.1735.

[4] Hwang, E., Yin, Y., Carenini, G., West, P., Shwartz, V. (2026). Infusing theory of mind into socially intelligent LLM agents. In Findings of the Association for Computational Linguistics: ACL 2026, pages 11327–11360. Association for Computational Linguistics. doi:10.18653/v1/2026.findings-acl.551.

[5] Li, M., Shi, X., Deng, Y. (2026). CoSToM: Causal-oriented steering for intrinsic theory-of-mind alignment in large language models. In ACL 2026, 64th Annual Meeting of the Association

for Computational Linguistics, pages 9302–9317. Association for Computational Linguistics. doi:10.18653/v1/2026.acl-long.421.

[6] Zhang, Z., Jin, C., Jia, M. Y., Zhang, S., Shu, T. (2025). AutoToM: Scaling model-based mental inference via automated agent modeling. In NeurIPS 2025, 39th Conference on Neural Information Processing Systems.

[7] Xiao, Y., Wang, J., Xu, Q., Song, C., Xu, C., Cheng, Y., Li, W., Liu, P. (2025). Towards dynamic theory of mind: Evaluating LLM adaptation to temporal evolution of human states. In ACL 2025, 63rd Annual Meeting of the Association for Computational Linguistics, pages 24036–24057. Association for Computational Linguistics.

[8] Zhou, X., Liu, J., Yerukola, A., Kim, H., Sap, M. (2025). Social world models. arXiv preprint arXiv:2509.00559.

[9] Yu, H., Zhao, Y., Lin, G., You, J. (2026). Building social world models with large language models. In ICML 2026, 43rd International Conference on Machine Learning, volume 306, pages 151719–151736. PMLR.

[10] Cross, L., Xiang, V., Bhatia, A., Yamins, D. L. K., Haber, N. (2025). Hypothetical Minds: Scaffolding theory of mind for multi-agent tasks with large language models. In ICLR 2025, International Conference on Learning Representations.

[11] Choi, Y., Li, C., Yang, Y., Jin, Z. (2025). Agent-to-agent theory of mind: Testing interlocutor awareness among large language models. In EMNLP 2025, Conference on Empirical Methods in Natural Language Processing, pages 28895–28928. Association for Computational Linguistics.

[12] Han, P., Liu, Z., You, J. (2026). ToMAP: Training opponent-aware LLM persuaders with theory of mind. In ICML 2026, 43rd International Conference on Machine Learning, volume 306. PMLR.

[13] Duncan, J. (2025). Construction and use of mental models: Organizing principles for the science of brain and mind. Neuropsychologia, volume 207, article 109062. doi:10.1016/j.neuropsychologia.2024.109062.

[14] Bonard, C. (2026). The belief–desire appraisal theory of emotion. Pacific Philosophical Quarterly. doi:10.1111/papq.70022.

[15] Yang, Y., Duan, Y., Huang, L., Zhu, Y., Bailo, F., Su, C., Chen, H. (2026). ScioMind: Cognitively grounded multi-agent social simulation with anchoring-based belief dynamics and dynamic profiles. arXiv preprint arXiv:2605.13725.

[16] Kolonin, A. (2026). Cognitive architecture and behavioral model based on social evidence and resource constraints. Brain Informatics, volume 13, article 8. doi:10.1186/s40708-026-00294-1.

[17] Hafner, D., Pasukonis, J., Ba, J., Lillicrap, T. (2025). Mastering diverse control tasks through world models. Nature, volume 640, pages 647–653. doi:10.1038/s41586-025-08744-2.

[18] Zhao, C., Zhang, R., Wang, J., Zhao, G., Niyato, D., Sun, G., Mao, S., Kim, D. I. (2025). World models for cognitive agents: Transforming edge intelligence in future networks. arXiv preprint arXiv:2506.00417.

[19] Bruce, J., Dennis, M. D., Edwards, A., Parker-Holder, J., Shi, Y., Hughes, E., Lai, M., et al. (2024). Genie: Generative interactive environments. In ICML 2024, 41st International Conference on Machine Learning, volume 235, pages 4603–4623. PMLR.

[20] Hafner, D., Lillicrap, T., Norouzi, M., Ba, J. (2021). Mastering Atari with discrete world models. In ICLR 2021, International Conference on Learning Representations.

[21] Richards, J., Wessel, M. (2024). What you need is what you get: Theory of mind for an LLM-based code understanding assistant. arXiv preprint arXiv:2408.04477.

[22] Amirizaniani, M., Martin, E., Sivachenko, M., Mashhadi, A., Shah, C. (2024). Do LLMs exhibit human-like reasoning? Evaluating theory of mind in LLMs for open-ended responses. arXiv preprint arXiv:2406.05659.

[23] Zhu, F., Simmons, R. (2024). Bootstrapping cognitive agents with a large language model. In AAAI-24, 38th AAAI Conference on Artificial Intelligence.

[24] van Duijn, M., van Dijk, B., Kouwenhoven, T., de Valk, W., Spruit, M., van der Putten, P. (2023). Theory of mind in large language models: Examining performance of 11 state-of-the-art models vs. children aged 7–10 on advanced tests. In CoNLL 2023, 27th Conference on Computational Natural Language Learning, pages 389–402. Association for Computational Linguistics.

[25] Xuanzang, trans. (7th century). Cheng Weishi Lun. 10 volumes. Taishō Tripiṭaka, volume 31, no. 1585.

[26] Kuiji (7th century). Cheng Weishi Lun Shuji. 10 volumes. Taishō Tripiṭaka, volume 43, no. 1830.

[27] Kuiji (7th century). Cheng Weishi Lun Zhangzhong Shuyao. 4 volumes. Taishō Tripiṭaka, volume 43, no. 1831.

[28] Lian, Z., Liu, R., Xu, K., et al. (2025). MER 2025: When affective computing meets large language models. In MM 2025, 33rd ACM International Conference on Multimedia, pages 13837–13842. ACM. doi:10.1145/3746027.3762007.

[29] Tak, A. N., Banayeeanzade, A., Bolourani, A., Kian, M., Jia, R., Gratch, J. (2025). Mechanistic interpretability of emotion inference in large language models. In Findings of the Association for Computational Linguistics: ACL 2025, pages 13090–13120. Association for Computational Linguistics. doi:10.18653/v1/2025.findings-acl.679.

[30] Cai, J., Zhang, X., Pan, Y., et al. (2026). An emotion recognition dataset using millimeter wave radar and physiological reference signals. Scientific Data, volume 13, article 820. doi:10.1038/s41597-026-07159-6.